\documentclass{article} 
\usepackage[table]{xcolor}
\usepackage{iclr2022_conference,times}
\definecolor{mydarkblue}{rgb}{0,0.08,0.45}
\usepackage[colorlinks=true,
    linkcolor=mydarkblue,
    citecolor=mydarkblue,
    filecolor=mydarkblue,
    urlcolor=mydarkblue]{hyperref}

\usepackage{amsmath,amsfonts,bm}

\def\eqref#1{equation~\ref{#1}}

\def\1{\bm{1}}

\DeclareMathAlphabet{\mathsfit}{\encodingdefault}{\sfdefault}{m}{sl}
\SetMathAlphabet{\mathsfit}{bold}{\encodingdefault}{\sfdefault}{bx}{n}

\usepackage{url}
\usepackage{graphicx}
\usepackage{multirow}
\usepackage{booktabs} 
\usepackage{tikz}
\usepackage{amsfonts}
\usepackage{amsmath}
\usepackage{amssymb}
\usepackage{nicefrac}
\usepackage{siunitx}
\usepackage{textcomp}
\usepackage{url}
\usepackage{color}
\usepackage{xcolor,colortbl}
\usepackage{xspace}
\usepackage{times}
\usepackage{textcomp}
\usepackage{lipsum}
\usepackage{pifont}
\usepackage{multirow}
\usepackage{aliascnt}
\usepackage{threeparttable}
\usepackage{wrapfig}
\usepackage{soul}
\usepackage{algorithm}
\usepackage{algpseudocode}

\usepackage{caption}
\usepackage{subcaption}

\newcommand{\xmark}{\ding{55}}%

\title{Generative Kernel Continual learning}

\author{Mohammad Mahdi Derakhshani\textsuperscript{1}, Xiantong Zhen\textsuperscript{1,2}, Ling Shao\textsuperscript{2}, Cees G. M. Snoek\textsuperscript{1} \\
\textsuperscript{1}AIM Lab, University of Amsterdam\\
\textsuperscript{2}Inception Institute of Artificial Intelligence\\
}

\usepackage{sidecap}

\iclrfinalcopy 
\begin{document}

\maketitle

\begin{abstract} 
Kernel continual learning by \citet{derakhshani2021kernel} has recently emerged as a strong continual learner due to its non-parametric ability to tackle task interference and catastrophic forgetting. Unfortunately its success comes at the expense of an explicit memory to store samples from past tasks, which hampers scalability to continual learning settings with a large number of tasks. In this paper, we introduce generative kernel continual learning, which explores and exploits the synergies between generative models and kernels for continual learning. The generative model is able to produce representative samples for kernel learning, which removes the dependence on memory in kernel continual learning. Moreover, as we replay only on the generative model, we avoid task interference while being computationally more efficient compared to previous methods that need replay on the entire model. We further introduce a supervised contrastive regularization, which enables our model to generate even more discriminative samples for better kernel-based classification performance. We conduct extensive experiments on three widely-used continual learning benchmarks that demonstrate the abilities and benefits of our contributions. Most notably, on the challenging SplitCIFAR100 benchmark, with just a simple linear kernel we obtain the same accuracy as kernel continual learning with variational random features for one tenth of the memory, or a 10.1\% accuracy gain for the same memory budget.
\end{abstract}

\section{Introduction}

Continual learning~\citep{ring1998child, lopez2017gradient, Goodfellow2013AnEI}, also known as lifelong learning, strives to continually learn to solve a sequence of non-stationary tasks. The continual learner is required to accommodate new information, while maximally maintaining the knowledge acquired in previous tasks so as to be still able to complete those tasks. This is a challenge for contemporary deep neural networks as they suffer from catastrophic forgetting when learning over non-stationary data \citep{McCloskey1989CatastrophicII}. 
%

Kernel continual learning by~\cite{derakhshani2021kernel}  recently emerged as a strong continual learner by combining the strengths of deep neural networks and kernel learners~\citep{scholkopf2001generalized, smola1998learning, rahimi2007random}. Kernel continual learning deploys a non-parametric classifier based on kernel ridge regression, which systematically avoids task interference and offers an effective way to deal with catastrophic forgetting. Moreover, by sharing the feature extraction and kernel inference networks, useful knowledge is transferred across tasks. Kernel continual learning performs well in both the task-aware and domain incremental learning scenarios, while being simple and efficient in terms of architecture and training time. 
%
%

Despite its appealing abilities, the success of kernel continual learning comes at the expense of an explicit memory that needs to maintain the data of all previously experienced tasks. It includes an episodic memory unit to store a subset of samples from the training data for each task, called the `coreset', from which a classifier learns based on kernel ridge regression. 
In order to perform well, a large memory is required to construct a satisfactory kernel, which causes computation and storage overhead when learning along with a growing number of tasks. In addition, the coreset is constructed by drawing samples uniformly from existing classes in the same task. This uniform sampling strategy is unlikely to provide the most representative and discriminative samples per task, potentially hurting accuracy. 

\begin{figure*}[t!]
\centering
\includegraphics[clip, trim=0.7cm 0cm 1cm 0cm, width=1\textwidth]{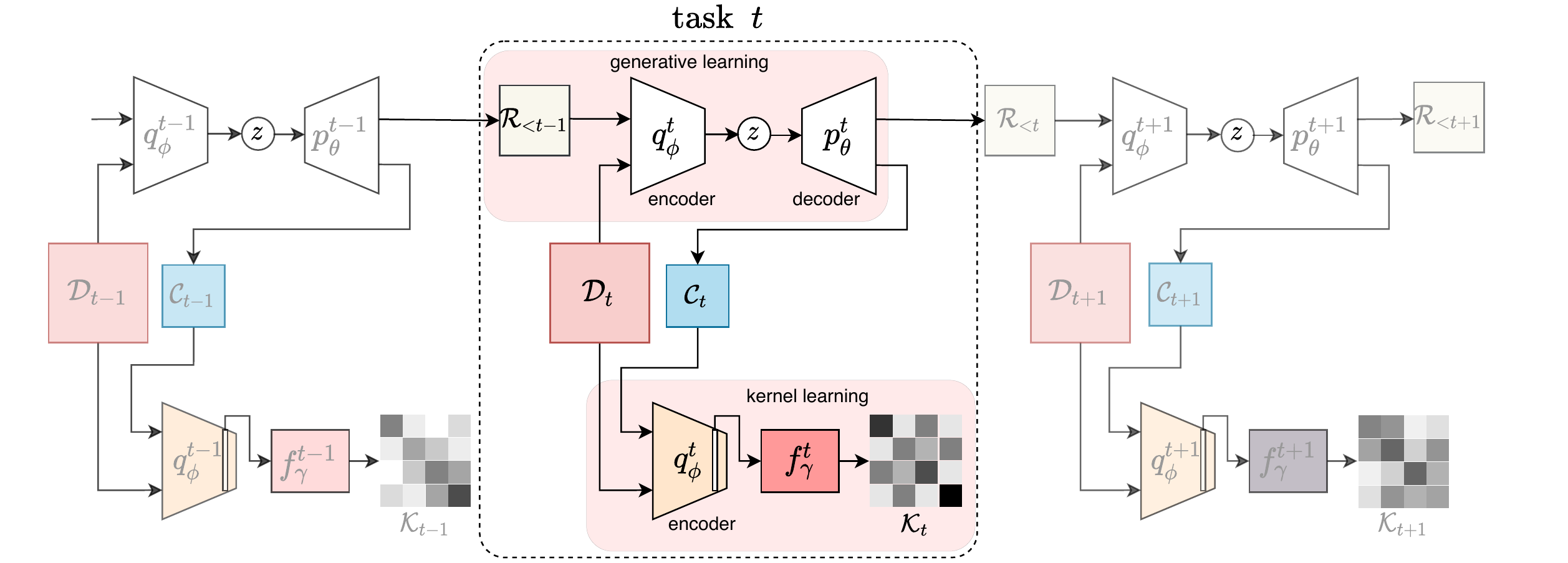}
\caption{\textbf{Overview of generative kernel continual learning}. 
The variational auto-encoder is adopted as a generative model, which learns the data distribution of task $t$ while doing replay over the data $\mathcal{R}_{<t}$ from previous tasks to avoid catastrophic forgetting. The trained decoder of the generative model is deployed to generate the coreset  $\mathcal{C}_t$. The kernel $\mathcal{K}_t$ for the current task $t$ is constructed based on the generated $\mathcal{C}_t$ and $\mathcal{D}_t$. The classifier for task $t$ is constructed based on kernel ridge regression using $\mathcal{K}_t$. 
The encoder network $q_{\phi}^t$,  parameterized by $\phi$, is shared and updated when training on the task sequence. 
The decoder network $p_{\theta}$, parameterized with $\theta$, adopts a gating mechanism that avoids task interference and minimizes forgetting.
The kernel network $f_{\gamma}^t$, parameterized by $\gamma$, infers a task-specific kernel, which is shared among tasks and trained end-to-end.
The shaded network $q_{\phi}^t$ is the trained encoder network from the generative model, which functions as a feature extractor to produce internal representations for kernel learning. 
}
\label{fig:demo}
\label{ckcl}
\end{figure*}

To alleviate the shortcomings of kernel continual learning while leveraging the merits, we introduce \textit{generative kernel continual learning}, a memory-less variant of kernel continual learning that replaces the episodic memory unit with a generative model based on the conditional variational auto-encoder \citep{van2020brain} that can learn task distributions sequentially.
This change allows the kernel continual learning method to easily increase the coreset size and draw samples from each task’s data distribution based on data point likelihoods. As a result, the kernel approximation is enhanced, even for small sample sizes per class. We make three contributions in this paper:  
\begin{enumerate}
    \item We propose generative kernel continual learning by synergizing the strengths of generative models and kernels. It removes the dependence on an explicit memory while still being able to tackle catastrophic forgetting and task interference.
    \item We introduce a supervised contrastive loss into the generative modelling, which increases the discriminability of the generated latent representations, further improving the model's performance in terms of accuracy and forgetting.
    \item We demonstrate generative kernel continual learning with the most simple linear kernel already outperforms the most advanced kernel continual learning variant with variational random features and sets a new state-of-the-art for both task-aware and domain incremental continual learning. 
\end{enumerate}
The schematic overview of the proposed generative kernel continual learning is depicted in Figure~\ref{ckcl} and detailed next.

\section{Generative Kernel Continual Learning}

A continual learning agent learns to solve a sequence of non-stationary tasks. Following the task setup in~\citep{mirzadeh2020understanding, chaudhry2020continual, Chaudhry2019OnTE}, we consider continual learning as a sequence of tasks $\{1, \cdots, t, \cdots, T\}$ that are proceeded one at a time, where $T$ is the total number of tasks. Each task $t$ is a classification problem with its own dataset $D_t{=}(\mathbf{x}_i^t, \mathbf{y}_i^t)_{i=1}^{N_t}$ where $\mathbf{x}_i^t$ is the $i$-th input vector, $\mathbf{y}_i^t$ is its corresponding output target, and $N_t$ refers to the number of input-output pairs regarding task $t$.

\subsection{Kernel Continual Learning} 
\label{lbl:preliminaries}

Kernel continual learning \citep{derakhshani2021kernel} deploys a non-parametric classifier based on kernel ridge regression, which does not require memory replay and systematically avoids task interference by adopting a task-specific classifier. 
Specifically, during training task $t$, dataset $D_t$ is split into two disjoint subsets $C_t$ and $D_t \setminus C_t$. As the model, a neural network $h_\theta$ is deployed that provides the feature vector $\psi(\mathbf{x}) = h_\theta(\mathbf{x}) \in \mathbb{R}^d$ as its output. $h_\theta(\cdot)$ is shared among all tasks, and $\theta$ denotes its parameters. To learn task $t$, a classifier $f_c^t$ is built upon the feature vectors extracted from $h_\theta(\cdot)$. In the probabilistic perspective, the predictive distribution is denoted as follows:
\begin{equation}
    \label{formula:kcl}
    \tilde{\mathbf{y}}^{\prime}=f_{c}^{t}\left(h_{\theta}\left(\mathbf{x}^{\prime}\right)\right)=\operatorname{softmax}\left(Y(\lambda I+\mathcal{K})^{-1} \tilde{K}\right),
\end{equation}
where $\mathcal{K}{=}\psi(X)^\top  \psi(X)$, $\tilde{K}{=}\psi(X)^\top  \psi(\mathbf{x}^{\prime})$, $\psi(X) \in \mathbb{R}^{d \times N_c}$ is the feature matrix that contains the feature vectors of the coreset samples, $Y$ is the one-hot representation of the target of the coreset samples, and $\psi(\mathbf{x}^{\prime})$ denotes the features extracted from a query sample of $D_t \setminus C_t$. Finally, $h_\theta$ is trained on the current task using the cross-entropy loss function. It is further supposed that the task boundary is known during training time and inference time, and $C_t$ is accumulated in the memory unit $\mathcal{M}$ to be used at inference time.


The performance of kernel continual learning is highly dependent on the quality of the coreset. 
In general, a coreset should satisfy two properties. It should be as small as possible but representative of the full dataset. Ideally, a model trained on the coreset performs as well as one trained on the full dataset \citep{borsos2020coresets}.
For kernel continual learning, \cite{derakhshani2021kernel} confirm these facts and show a positive correlation exists between the coreset size and the continual learning model accuracy. 
That is, a larger, and as a result more representative, coreset leads to a better approximation of kernel terms $\mathcal{K}$ and $\tilde{K}$ in Eq.~(\ref{formula:kcl}). Moreover, kernel continual learning also adopt variational random features to learn task-specific kernels, which also contributes greatly to the overall performance. In this paper, we work with predefined kernels without inducing learnable parameters in kernels.

\subsection{Generative Kernel Learning} \label{lbl:coreset_seclection_phase}
In kernel continual learning, the coreset is constructed by selecting samples from the training set of each task~\citep{derakhshani2021kernel}. However, this coreset selection mechanism has two complications. First, it relies on a uniform sampling strategy to draw samples from the task dataset $D_t$, with no guarantee that the selected samples are sufficiently representative of the data in each task.
Second, the coreset size is usually bounded by memory storage constraints, which limits its capacity and leads to a drop in model performance as learning proceeds with an increasing number of tasks.

In order to alleviate the shortcomings while enjoying the merits of kernel continual learning, we propose to replace the fixed size memory unit with a generative model. The generative model removes the dependence on the memory unit in kernel continual learning by generating samples to construct kernels for each task. As an extra benefit, by using generative modelling, the model is allowed to flexibly generate as many samples as needed, without being restricted by memory size in kernel continual learning. To be more specific, we resort to generative replay on internal representations~\citep{van2020brain} based on variational auto-encoders~\citep{kingma2013auto}, as it allows kernel continual learning to draw samples proportional to their likelihood values in the coreset selection phase.

\begin{figure}
  \begin{center}
    \includegraphics[width=.95\textwidth]{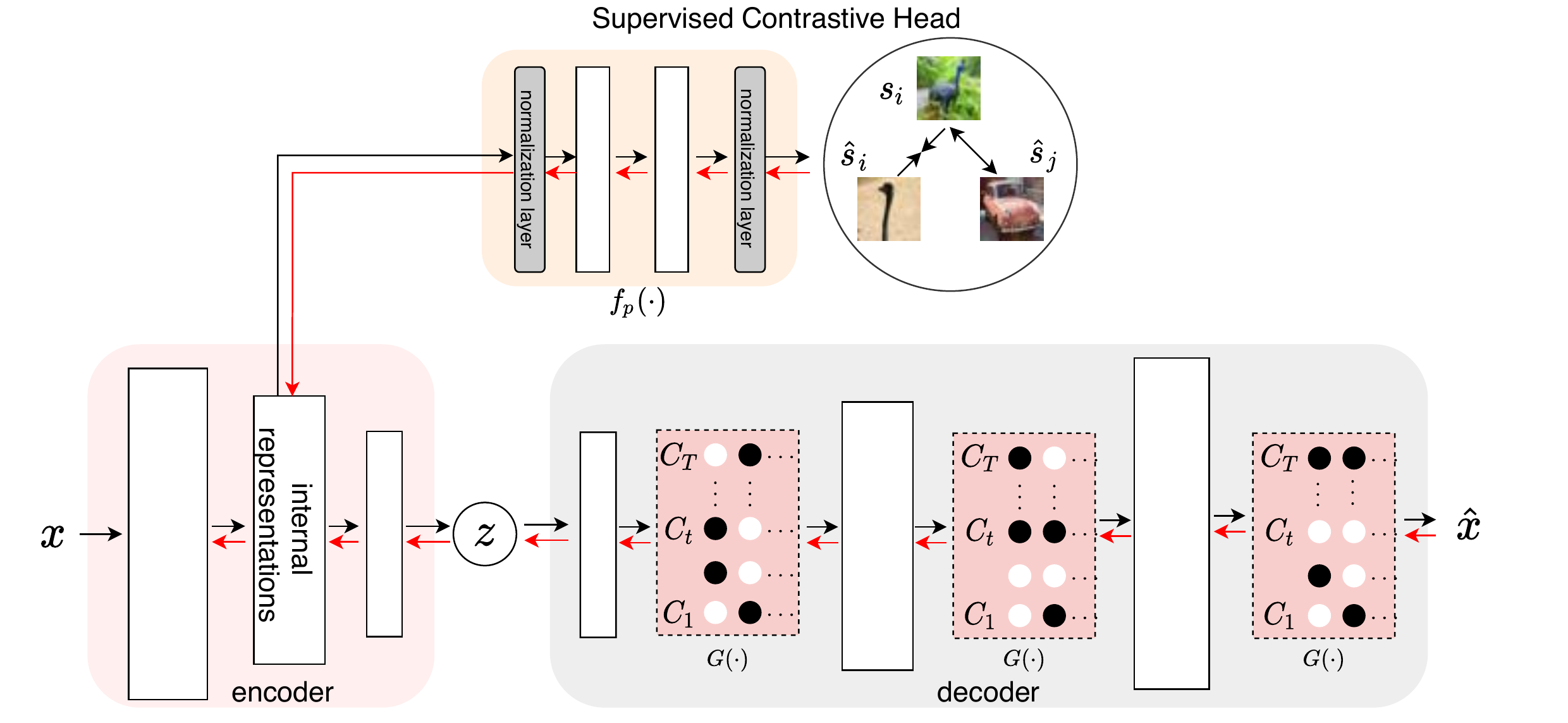}
  \end{center}
  \vspace{-2mm}
  \caption{\textbf{Generative learning by a variational auto-encoder with supervised contrastive regularization}. 
  The internal representations from the intermediate layers of the encoder are used to kernel learning. A projection network $f_p(\cdot)$ is applied to the internal representations, which are fed into the supervised contrastive loss function for discriminative sample generation. In the decoder network, as in~\citep{van2020brain}, we adopt a gating function $G(\cdot)$ in red blocks to avoid task interference. Black and red arrows show the forward and backward path respectively. 
  %
  }
\label{fig:scl}
\vspace{-4mm}
\end{figure}

%
The generative model with a variational auto-encoder is shown in Figure~\ref{fig:scl}. Similar to \citep{von2019continual, wortsman2020supermasks}, the decoder applies a gating mechanism which allows the model to adjust its decoder output per task or domain to minimize the task interference problem. More specifically, each layer consists of several gates in the decoder network. They are binary, randomly pre-initialized and fixed during training, and they are multiplied to the output of each layer as \citep{van2020brain}. Gates are defined according to the continual learning scenarios. For the task incremental learning, the number of gates in each layer is equivalent to the number of classes while for the domain incremental learning, it is equal to the number of domains in the given benchmark.
Furthermore, in contrast to the vanilla variational auto-encoder \citep{kingma2013auto} that uses a factorized normal distribution as its prior distribution, we exploit a learnable mixture of Gaussian distributions, in which each component corresponds to one class. Treating the prior distribution in this manner helps capture the aggregated posterior objective better \citep{tomczak2018vae}.

\paragraph{Training objective}
Training of generative kernel continual learning involves two successive steps. In the first step, we train the variational auto-encoder to be capable of generating samples regarding the current tasks $t$ as well as samples from previous tasks ($\mathcal{R}_{<t}$). Then, in the second step, using the decoder of the trained generative model, we produce the coreset for the current task $t$. Next, we put both $C_t$ and $D_t$ through the encoder network $q_{\phi}$ and obtain their corresponding internal representations for learning kernels. These internal representations are decoupled from the model forward path and fed into the kernel network $f_\gamma$ for construct classifiers based on kernel ridge regression. Finally, we train the classifiers on top of the features extracted from the kernel network as defined in Section \ref{lbl:preliminaries}. 

During training the variational auto-encoder, we receive in each iteration the current task data $(\mathbf{x}_i, \mathbf{y}_i, \mathbf{t}_i)$ and the replayed data from previous tasks $(\mathbf{x}_j, \mathbf{y}_j, \mathbf{t}_j)$ where $\mathbf{x}$, $\mathbf{y}$ and $\mathbf{t}$ are the input vector, output target and the task identifier.  Both input vectors $\mathbf{x_i}$ and $\mathbf{x_j}$ are fed in the encoder network and their corresponding posterior distributions $\mathbf{z}_i \sim q_{\phi}(\mathbf{z}|\mathbf{x}_i)$ and $\mathbf{z}_j \sim q_{\phi}(\mathbf{z}|\mathbf{x}_j)$ are estimated. Then, we reconstruct the two input vectors using the decoder network as $\mathbf{x}_i \sim p_{\theta}(\mathbf{x} | \mathbf{z}_i, \mathbf{y}_i, \mathbf{t}_i)$ and $\mathbf{x}_j \sim p_{\theta}(\mathbf{x} | \mathbf{z}_j, \mathbf{y}_j, \mathbf{t}_j)$. Finally, we minimize the following overall objective function per iteration:
\begin{equation}
\begin{aligned}
\mathcal{L}(\theta, \phi) = \underbrace{\mathcal{L}(\theta, \phi ; \mathbf{x}_i, \mathbf{y}_i, \mathbf{t}_i,  \mathbf{z}_i)}_\text{current task data} + \underbrace{\mathcal{L}(\theta, \phi ; \mathbf{x}_j, \mathbf{y}_j, \mathbf{t}_j,  \mathbf{z}_j)}_\text{replayed data},
\end{aligned}
\label{formula:objective}
\end{equation}
where each term in Eq.~(\ref{formula:objective}) is the following evidence lower-bound objective function:
\begin{equation}
\begin{aligned}
\mathcal{L}(\theta, \phi ; \mathbf{x}, \mathbf{y}, \mathbf{t},  \mathbf{z})=\mathbb{E}_{q_{\phi}(\mathbf{z} | \mathbf{x})}\left[\log p_{\theta}(\mathbf{x} | \mathbf{z}, \mathbf{y}, \mathbf{t})\right]-D_{\rm{KL}}\left(q_{\phi}(\mathbf{z} | \mathbf{x}) \| p(\mathbf{z} | \mathbf{y})\right).
\end{aligned}
\label{formula:elbo}
\end{equation}
In Eq.~(\ref{formula:elbo}), we define $p(\mathbf{z} | \mathbf{y}){=}\mathcal{N}(\mathbf{z} |\mu_y, \sigma_y I)$  as the prior distribution for class $y$, which is a factorized normal distribution and its parameters $\mu_y$ and $\sigma_y$ are trainable. 

Our formulation for generative kernel continual learning encourages the model to produce pseudo-samples from observed tasks and generate the coreset accordingly. However, its encoder network $q_{\phi}(\mathbf{z} | \mathbf{x})$, which is used as the feature extractor for kernel continual learning, would not necessarily be able to provide sufficiently discriminative samples. To alleviate this problem, we propose to exploit supervised contrastive learning \citep{khosla2020supervised} in generative kernel continual learning to further improve the discrimination of the latent representation. 

\subsection{Supervised Contrastive Regularization}
\label{lbl:contrastive_loss}
In order to achieve better classification performance, the generated samples are required to be discriminative. To this end, we further introduce a supervised contrastive regularizer~\citep{khosla2020supervised} into the optimization of the variational auto-encoders. We impose the supervised contrastive term on the internal representations that are used for kernel learning. 
Generally, in contrastive learning methods~\citep{chen2020simple, he2020momentum, chen2021exploring}, a data augmentation operation is applied over the current batch data to obtain two different copies of the input batch. Both copies are fed through an encoder network, and a normalized embedding vector is estimated. At training time, these normalized vectors are further forward propagated through a projection network that is thrown away at inference time.

We adopt a similar contrastive learning strategy. Given data of the current task $(\mathbf{x}_i, \mathbf{y}_i, \mathbf{t}_i)$ and the data to replay from previous tasks $(\mathbf{x}_j, \mathbf{y}_j, \mathbf{t}_j)$, we stack them together to obtain a new set: $\mathbf{x}{=}[\mathbf{x}_i; \mathbf{x}_j]$ with the associated target labels $\mathbf{y}{=}[\mathbf{y}_i; \mathbf{y}_j]$. We consider $(\mathbf{x}, \mathbf{y})$ as the current batch. Then, we put $\mathbf{x}$ into the conditional variational auto-encoder as the data augmentation operation to generate a random augmentation of the input vector as  $\hat{\mathbf{x}}=\operatorname{Aug(\mathbf{x})}=\operatorname{Decoder}(\operatorname{Encoder}(\mathbf{x}))$. 
As for the encoder network, we utilize $q_{\phi}(\mathbf{z} |\mathbf{x})$ that maps the $\mathbf{x}$ and $\hat{\mathbf{x}}$ into an internal representation vector $r \in \mathbb{R}^k$. Then, this representation vector is normalized to the unit hypersphere to enhance the top-1 accuracy as done in \citep{khosla2020supervised}. In the next step, we forward the normalized $r$ through a fully connected neural network, namely, the projection network $f_p(\cdot)$, and once more normalize the projection output $\mathbf{s} \in \mathbb{R}^v$. Finally, we compute the following supervised contrastive loss function and optimize it in conjunction with the main loss function introduced in Eq.~(\ref{formula:objective}):
\begin{equation}
\begin{aligned}
\mathcal{L}^{s}=\sum_{i \in I} \mathcal{L}_{i}^{s},
\end{aligned}
\label{formula:contrastive}
\end{equation}
\begin{equation}
\begin{aligned}
 \mathcal{L}_{i}^s= \frac{-1}{|P(i)|} \sum_{p \in P(i)} \log \frac{\exp \left(\boldsymbol{s}_{i} \cdot \boldsymbol{s}_{p} / \tau\right)}{\sum_{a \in A(i)} \exp \left(\boldsymbol{s}_{i} \cdot \boldsymbol{s}_{a} / \tau\right)},
\end{aligned}
\label{formula:contrastive_each_sample}
\end{equation}
where $i$ is the index of all existing samples $\{1 \cdot\cdot\cdot 2N\}$ including samples in both $\mathbf{x}$ and $\hat{\mathbf{x}}$, $A(i){=}I \setminus \{i\}$, $P(i){=} \{p \in A(i) : y_p {=} y_i \}$ is the set of all indices having the same class label as $i$, $|P(i)|$ is the set cardinality, and $\tau$ is the temperature parameter. 


\section{Experiments}
\subsection{Benchmarks, Metrics \& Details}
\label{lbl:experiments_benchmarks}
\paragraph{Three benchmarks} 
We evaluate generative kernel continual learning for two well-established scenarios: \textit{domain incremental learning} and \textit{task incremental learning}. For the domain incremental learning scenario, we rely on PermutedMNIST~\citep{kirkpatrick2017overcoming} and RotatedMNIST~\citep{mirzadeh2020understanding}, and for the task of incremental learning, we report on SplitCIFAR100~\citep{zenke2017continual}. Each PermutedMNIST task is produced by a random permutation of the image pixels, such that this permutation remains fixed within the same task. Each RotatedMNIST task is composed by randomly rotating the input images by a degree between $0$ and $180$, such that the rotation degree is the same within a task. 
SplitCIFAR100 is a continual variant of CIFAR100 where each task represents the data from 5 out of 100 random classes.

\paragraph{Two metrics} We report the continual learning performance using average accuracy $\mathrm{a}_{t}$ and average forgetting $F$, following \citep{titsias2019functional,mirzadeh2020understanding,chaudhry2020continual,derakhshani2021kernel}. The average accuracy of a model when the training of task $t$ is finished equals:
\begin{equation}
\begin{aligned}
\mathrm{A}_{t}=\frac{1}{t} \sum_{i=1}^{t} \mathrm{a}_{t, i},
\end{aligned}
\label{formula:average_accuracy}
\end{equation}
where $\mathrm{a}_{t, i}$ refers to the accuracy of the model on task $i$ after being trained on task $t$. Average forgetting measures the drop in model performance between the highest accuracy and the last accuracy of each task when the continual learning model training is finished. It is calculated as:
\begin{equation}
\begin{aligned}
\mathrm{F}=\frac{1}{T-1} \sum_{i=1}^{T-1} \max _{1, \ldots, \mathrm{T}-1}\left(\mathrm{a}_{\mathrm{t}, \mathrm{i}}-\mathrm{a}_{\mathrm{T}, \mathrm{i}}\right).
\end{aligned}
\label{formula:average_forgetting}
\end{equation}

\paragraph{Implementation details}
Generative kernel continual learning is implemented with a deep neural network composed of three networks: a variational auto-encoder, which itself has an encoder network $q_{\phi}(\mathbf{z} |\mathbf{x})$ and a decoder network $p_{\theta}(\mathbf{x} |\mathbf{z}, \mathbf{y}, \mathbf{t})$, a projection network $f_p(\cdot)$, and a fully-connected network $f_{\gamma}$ for kernel learning. For PermutedMNIST and RotatedMNIST, the encoder $q_{\phi}(\mathbf{z} |\mathbf{x})$ consists of two fully connected layers, each having $2000$ neurons. For SplitCIFAR100, similar to \cite{van2020brain}, a pre-trained convolutional neural network precedes the encoder $q_{\phi}(\mathbf{z} |\mathbf{x})$. We pretrain this convolutional neural network on the classification task of CIFAR10 and only use its first five convolutional layers. For PermutedMNIST and RotatedMNIST, the pretrained network is an identity mapping function. The encoder $q_{\phi}(\mathbf{z} |\mathbf{x})$ consists of two fully connected layers, each of which has $256$ neurons. For all benchmarks, the latent distribution of the  variational auto-encoder consists of one fully connected layer of $100$ neurons parameterizing the mean and log standard deviation. The decoder $p_{\theta}(\mathbf{x} |\mathbf{z}, \mathbf{y}, \mathbf{t})$ is exactly the transpose of the encoder $q_{\phi}(\mathbf{z} |\mathbf{x})$. 
With regard to projection network $f_p(\cdot)$, we use one fully-connected layer with hidden size of $196$ neurons for PermutedMNIST and RotatedMNIST, and  $256$ neurons for SplitCIFAR100. Finally,  $f_{\gamma}$ includes two fully connected layers with $512$ neurons, followed by a dropout layer for all benchmarks. 
We implement generative kernel continual learning in Pytorch and the code will be released.

In order to make fair comparisons with other models, we train $f_{\gamma}$ with the same hyperparameters as \citep{derakhshani2021kernel}.
Unless otherwise stated, all models in our experiments rely on generative kernel continual learning with supervised contrasitve regularization, trained over $20$ sequential tasks on 5 different random seeds. The coreset sizes for both training and inference are similar and all experiments exploit a simple linear kernel. We train the conditional generative model for $2000$ iterations on PermutedMNIST and RotatedMNIST and $300$ iterations on SplitCIFAR100 per task. Following  \citep{derakhshani2021kernel}, the kernel network $f_\gamma$ is trained for one epoch per task and the batch size is set to $10$ (see appendix \ref{lbl:hyper} for more details).

\subsection{Results}

\begin{SCfigure}[][t]
\centering
\includegraphics[width=0.67\textwidth]{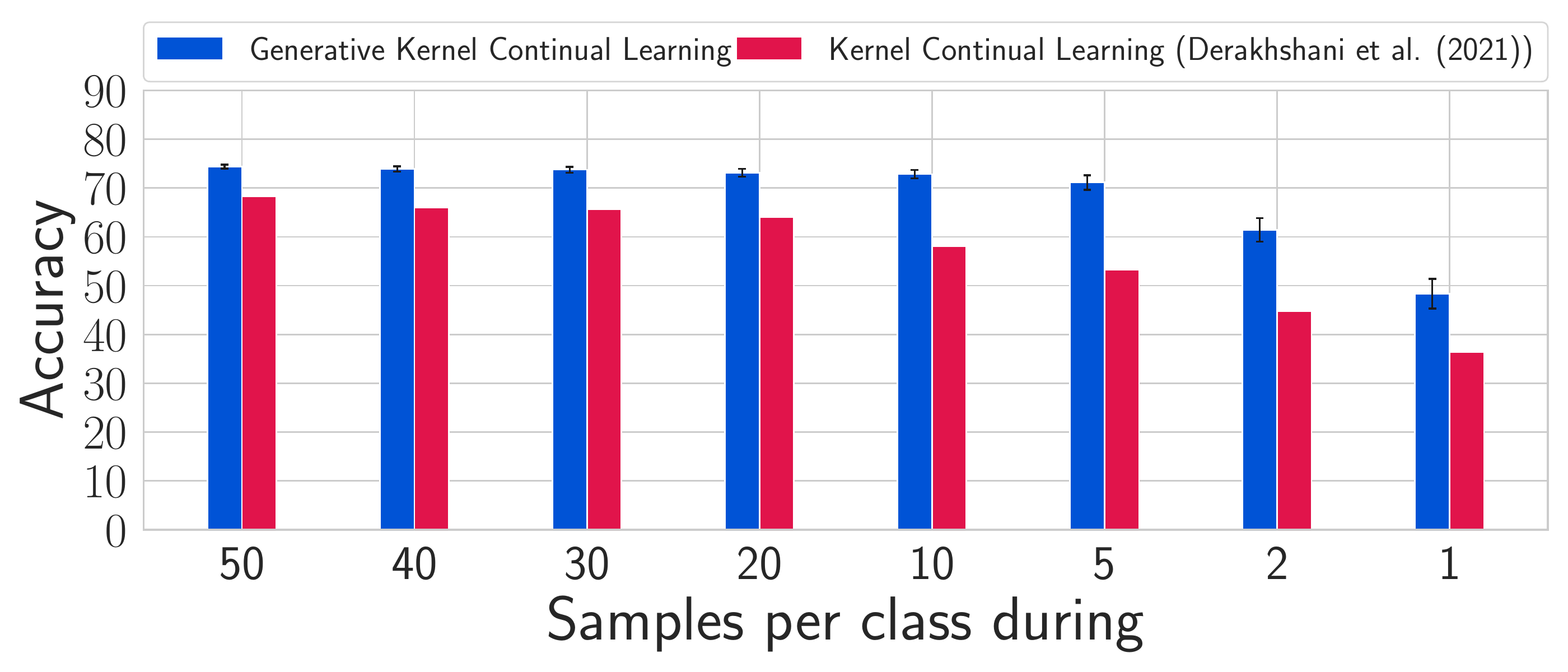}
\caption{\textbf{Generative coreset vs. uniform coreset} in kernel continual learning on SplitCIFAR100. Generative kernel continual learning needs about 10 times smaller coreset size for same accuracy, highlighting the expressiveness of the generated coreset. 
}
\label{fig:coreset_size_ablation}
\end{SCfigure}

\paragraph{Generative coreset vs. uniform coreset}
The number of samples per class as well as the quality of each sample in the coreset are considered to be two pivotal factors in obtaining good performance in kernel continual learning. To understand them better, we ablate the influence of the uniform coreset in kernel continual learning with the generative coreset in our approach using a varying number of samples per class in the coreset.
As Figure \ref{fig:coreset_size_ablation} reveals, generative kernel continual learning performs better than kernel continual learning, independent of the number of samples per class.  
Interestingly, the difference between the average accuracy of the two methods increases in favor of our generative approach as the number of samples per class are decreased. Generative kernel continual learning only needs to generate $2$ samples per class to achieve a similar accuracy as kernel continual learning with $20$ samples.
This shows that the generative model endows kernel continual learning with more representative and expressive coreset samples. The benefit increases when only a small number of samples per class can be stored. A smaller number of samples reduces the number of operations needed to compute kernels in equation \ref{formula:kcl}, and as a result, it reduces the run time of generative kernel continual learning during both training and inference.

\begin{SCfigure}[][t]
\centering
\includegraphics[width=0.66\textwidth]{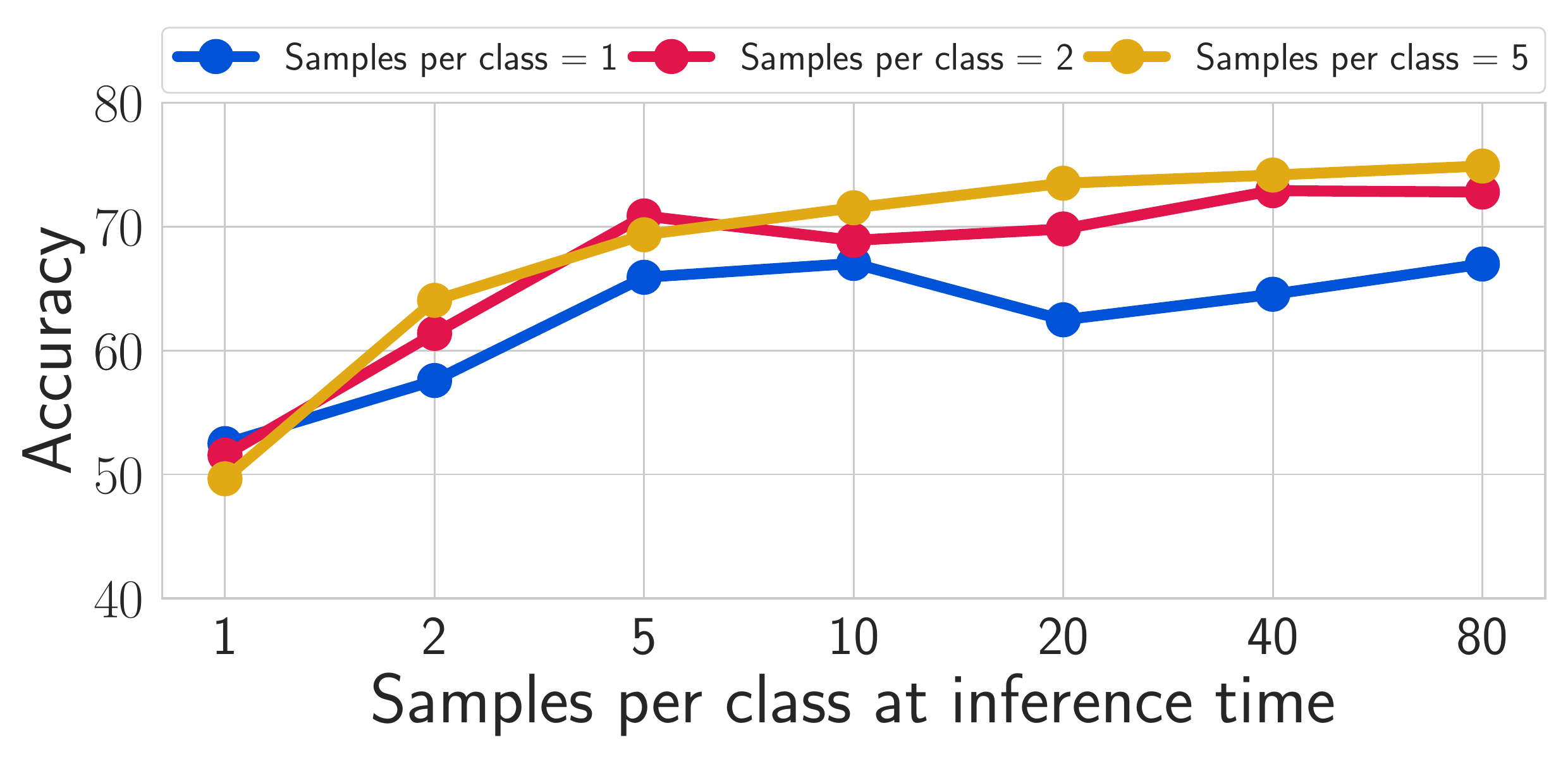}
\vspace{-4mm}
\caption{\textbf{Varying coreset size at inference time.} Average accuracy over 20 sequential tasks on SplitCIFAR100 when the coreset size differs between training (legend) and inference time. Increasing the coreset size during inference increases average accuracy considerably.}
\label{fig:variable_coreset_in_train_and_infernce}
\end{SCfigure}

\paragraph{Varying coreset size at  inference time}
Due to the generative nature of our proposal, the coreset size no longer needs to be the same during training and inference time, as in kernel continual learning. We run an ablation to show how average accuracy of our model changes by fixing the coreset size during training and varying the coreset size during inference. To do so, we train generative kernel continual learning with three different coreset sizes where we generate $1$, $2$ and $5$ samples per  class as depicted in Figure \ref{fig:variable_coreset_in_train_and_infernce}. For each coreset, we evaluate the trained model on seven different coreset sizes at inference time. 
As expected, model accuracy decreases when the coreset size at inference time is smaller than the size used during training time. That is to say, as we decrease the coreset size during inference below the coreset size during training time, the kernel network $f_\gamma$ receives less information than it expects to observe, and consequently it does poorly on the kernel approximation.
However, enlarging the coreset size at inference time, beyond the size used during training, always leads to an improvement. This behaviour shows the potency of kernel network $f_\gamma$ to utilize the extra information to enhance the model accuracy.
Hence, we suggest to train generative kernel continual learning with a smaller coreset size to reduce the training time and evaluate the same model on a larger coreset size for better model accuracy.

\paragraph{Influence of supervised contrastive regularization}
We introduce a supervised contrastive regularizer in generative kernel continual learning to enhance the discriminative power of the encoder's latent representation. To measure its benefit, we ablate generative kernel continual learning with and without the supervised contrastive regularization term in left Table \ref{tab:ssl_vs_not_ssl}, where we report each experiment on five different random seeds. As expected, adding the supervised contrastive regularizer leads to a consistent improvement in accuracy on all three benchmarks. In right Table \ref{tab:ssl_vs_not_ssl}, we further experiment with temperature parameter $\tau$ in Eq.  (\ref{formula:contrastive}) on SplitCIFAR100 over $1$ random seed, where hyperparameter $\tau$ adjusts the compactness and concentration of positive and negative samples in contrastive learning. It shows that our method is hardly sensitive to this hyperparameter and we consider fine-tuning optional.

\begin{table}[b]
    \caption{\textbf{Influence of supervised contrastive regularization} on generative kernel continual learning over PermutedMNIST, RotatedMNIST and SplitCIFAR100. Left: the contrastive regularizer further enhances continual learning average accuracy. Right: optimizing temperature parameter $\tau$ in Eq. (\ref{formula:contrastive}) results in a small gain on SplitCIFAR100.}
    \label{tab:ssl_vs_not_ssl}
    \begin{minipage}{.55\linewidth}
      \centering
        \resizebox{1\columnwidth}{!}{%
        \begin{tabular}{lccc}
        \toprule
        & \textbf{RotatedMNIST} & \textbf{PermutedMNIST} & \textbf{SplitCIFAR100} \\ 
        \midrule
        Without regularizer & 79.80\scriptsize{$\pm$2.02} & 88.23\scriptsize{$\pm$0.69} & 71.87\scriptsize{$\pm$0.56} \\
        \midrule
        With regularizer & \textbf{82.48\scriptsize{$\pm$1.33}} & \textbf{89.23\scriptsize{$\pm$0.38}} & \textbf{72.79\scriptsize{$\pm$0.68}}  \\
        \bottomrule
        \end{tabular}%
}
    \end{minipage}%
    \begin{minipage}{.45\linewidth}
      \centering
        \resizebox{0.9\columnwidth}{!}{%
        \begin{tabular}{@{}lccccc@{}}
        \toprule
        & \multicolumn{5}{c}{\textbf{SplitCIFAR100}}\\
        \midrule
        Temperature ($\tau$) & 0.02 & 0.04 & 0.08 & 0.1 & 1 \\ 
        \midrule
        Average Accuracy  & 73.70 & \textbf{74.25} & 73.70 & 73.78 & 74.10 \\
        \bottomrule
        \end{tabular}%
        }
    \end{minipage} 
\end{table}

\paragraph{Ability to generate multiple kernel types} To determine how the kernel type changes the generative kernel continual learning accuracy, we train the proposed method with three different kernel types: linear, radial basis function and polynomial kernels, on all three benchmarks. To provide fair comparisons, we keep all hyperparameters fixed, and only change the kernel type. Table \ref{tab:kernel_type} shows the results. Generative kernel continual learning provides promising results, independent of the kernel type. As could be expected, the best kernel varies per benchmark. The linear kernel and radial basis function kernels perform better for RotatedMNIST and PermutedMNIST, while for SplitCIFAR100 the polynomial kernel provides best average accuracy. By and large, a simple linear kernel is a good baseline, but the kernel type could be taken into account during hyperparameter optimization when accuracy is critical.

\begin{table}[t!]
	\renewcommand\arraystretch{1}
	\begin{minipage}{0.3\linewidth}
    	\caption{\textbf{Ability to generate multiple kernel types.} Results are promising, independent of kernel type. }
    	\label{tab:kernel_type}	
	\end{minipage}
    \hspace{0.02\linewidth}
    \begin{minipage}{0.66\linewidth}
    \vspace{-2mm}
        \centering
        \resizebox{1\columnwidth}{!}{%
            \begin{tabular}{@{}lccc@{}}
            \toprule
            & \textbf{RotatedMNIST} & \textbf{PermutedMNIST} & \textbf{SplitCIFAR100} \\ 
            \midrule
            Linear & \textbf{82.48} & 89.23 & 72.79 \\
            Polynomial & 82.08 & 89.23 & \textbf{75.33}  \\
            Radial Basis Function & 81.42 & \textbf{89.59} & 72.68  \\
            \bottomrule
            \end{tabular}%
        }
	\end{minipage}
\end{table}

\begin{figure*}[t]
\centering
\includegraphics[width=1\textwidth]{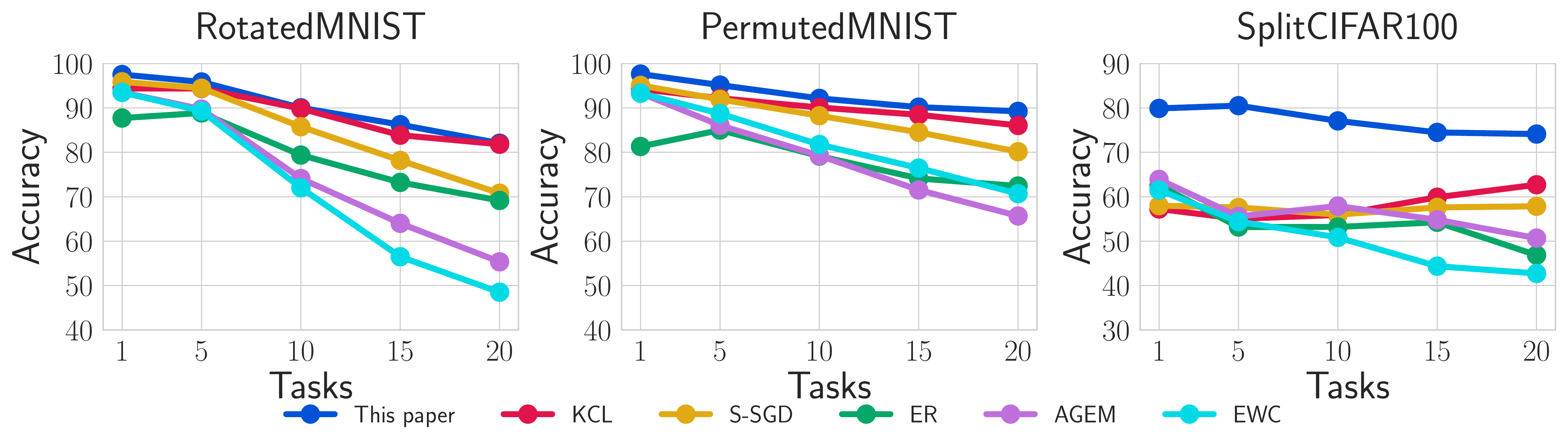}
\caption{\textbf{Comparison with state-of-the-art} over 20 consecutive
tasks, in terms of average accuracy. Our model consistently outperforms alternatives on all three benchmarks, especially on the more challenging SplitCFIAR100.
}
\label{fig:compare_all}
\vspace{-3mm}
\end{figure*}

\paragraph{Comparison with state-of-the-art}
To compare our proposed method against the state-of-the-art, we report average accuracy and average forgetting of generative kernel cotinual learning with supervised contrastive regularization for $20$ sequential tasks on 5 different random seeds over three well-established continual learning benchmarks in Table \ref{tab:compare-20tasks}. 
To provide a fair comparison with \cite{derakhshani2021kernel}, the coreset size at both inference and training time is similar and equal to $20$ for all benchmarks. Moreover, we exploit the simple linear kernel for all benchmarks. The temperature hyperparameter $\tau$ is $0.08$ for PermutedMNIST and RotatedMNIST and is $0.04$ for SplitCIFAR100. As shown, generative kernel continual learning outperforms all methods, setting a new state-of-the-art on PermutedMNIST, RotatedMNIST and SplitCIFAR100. In Figure \ref{fig:compare_all}, we visualize and compare our proposed model with alternative methods in terms of running average accuracy for $20$ sequential tasks. It can be seen that generative kernel continual learning performs consistently better than other methods on PermutedMNIST, RotatedMNIST and, especially, on SplitCIFAR100. We attribute the $10.1\%$ accuracy improvement on SplitCIFAR100 to the quality of the coreset samples provided by our generative model. For SplitCIFAR100, we also train generative kernel continual learning with a polynomial kernel, a coreset size of 5 samples per  class during training, and a coreset size of 80 samples per class during inference. We keep all other hyperparameters fixed. In this setting, we obtain an average accuracy of \textbf{76.5\scriptsize{$\pm$0.35}} and an average forgetting of \textbf{0.03\scriptsize{$\pm$0.00}}. 


\begin{table*}[t]
\centering
\caption{\textbf{Comparison with state-of-the-art.} Results for other methods are adopted from~\citet{derakhshani2021kernel}. Column \textit{unit} indicates whether methods exploit a memory unit $\mathcal{M}$ or a generative model $\mathcal{G}$. Generative kernel continual learning yields competitive results on PermutedMNIST and RotatedMNIST and outperforms alternatives on SplitCIFAR100 by a large margin. 
}
\vspace{-2mm}
\resizebox{1\linewidth}{!}{%
\begin{tabular}{lcccccccc}
\toprule
\multirow{2}{*}{\textbf{Method}} & 
\multirow{2}{*}{\textbf{Unit}} & 
\multicolumn{2}{c}{\textbf{Permuted MNIST}} & 
\multicolumn{2}{c}{\textbf{Rotated MNIST}} & 
\multicolumn{2}{c}{\textbf{Split CIFAR100}} \\  \cmidrule(lr){3-4} \cmidrule(lr){5-6} \cmidrule(lr){7-8} 
 &  & Accuracy & Forgetting & Accuracy & Forgetting & Accuracy & Forgetting\\ \midrule
Naive-SGD \citep{mirzadeh2020understanding} & \xmark  & 44.4\scriptsize{$\pm$2.46} & 0.53\scriptsize{$\pm$0.03} & 46.3\scriptsize{$\pm$1.37} & 0.52\scriptsize{$\pm$0.01} & 40.4\scriptsize{ $\pm$2.83} & 0.31\scriptsize{$\pm$0.02} \\
EWC \citep{kirkpatrick2017overcoming} & \xmark  & 70.7\scriptsize{$\pm$1.74} & 0.23\scriptsize{$\pm$0.01} & 48.5\scriptsize{$\pm$1.24} & 0.48\scriptsize{$\pm$0.01} & 42.7\scriptsize{$\pm$1.89} & 0.28\scriptsize{$\pm$0.03} \\
AGEM \citep{AGEM} & $\mathcal{M}$   & 65.7\scriptsize{$\pm$0.51} & 0.29\scriptsize{$\pm$0.01} & 55.3\scriptsize{$\pm$1.47} & 0.42\scriptsize{$\pm$0.01} & 50.7\scriptsize{$\pm$2.32} & 0.19\scriptsize{$\pm$0.04}\\
ER-Reservoir \citep{Chaudhry2019OnTE} & $\mathcal{M}$   & 72.4\scriptsize{$\pm$0.42} & 0.16\scriptsize{$\pm$0.01} & 69.2\scriptsize{$\pm$1.10} & 0.21\scriptsize{$\pm$0.01} & 46.9\scriptsize{$\pm$0.76} & 0.21\scriptsize{$\pm$0.03}\\
Stable SGD \citep{mirzadeh2020understanding} & \xmark  &  80.1\scriptsize{$\pm$0.51} & 0.09\scriptsize{$\pm$0.01} & 70.8\scriptsize{$\pm$0.78} & 0.10\scriptsize{$\pm$0.02} & 59.9\scriptsize{$\pm$1.81} & 0.08\scriptsize{$\pm$0.01} \\
Kernel Continual Learning \citep{derakhshani2021kernel} & $\mathcal{M}$  & 85.5\scriptsize{$\pm$0.78} & \textbf{0.02\scriptsize{$\pm$0.00}} & 81.8\scriptsize{$\pm$0.60} & 0.01\scriptsize{$\pm$0.00} & 62.7\scriptsize{$\pm$0.89} & 0.06\scriptsize{$\pm$0.01}\\
\textit{\textbf{Generative Kernel Continual Learning}} & $\mathcal{G}$  & \textbf{89.2\scriptsize{$\pm$0.44}} & 0.08\scriptsize{$\pm$0.00} & \textbf{82.4\scriptsize{$\pm$1.33}} & \textbf{0.01\scriptsize{$\pm$0.01}} & \textbf{72.8\scriptsize{$\pm$0.68}} & \textbf{0.04\scriptsize{$\pm$0.00}}\\
\bottomrule
\end{tabular}%
}
%
\label{tab:compare-20tasks}
\vspace{-2mm}
\end{table*}

\section{related work}
Following \cite{Lange2019ContinualLA}, we divide continual learning methodologies into three different categories. We have regularisation-based methods \citep{kirkpatrick2017overcoming, MAS, lee2017overcoming, zenke2017continual, kolouri2019attention} that regularize the neural network parameters to not change drastically from those learned on previous tasks. This goal is accomplished by estimating a penalty term for each parameter of the network using the Fisher information matrix \cite{kirkpatrick2017overcoming}, the gradient magnitude of each parameter \cite{MAS}, or by sequential Bayesian inference \cite{nguyen2017variational}. 
Recently, \cite{pmlr-v139-kapoor21b} proposes to exploit a variational auto-regressive Gaussian processes to improve the posterior distribution of sequential Bayesian inference methods due to the sequential nature of continual learning data. In general, these methods encourage the continual learning model to prioritize preserving old knowledge rather than absorbing new information from new tasks. Hence, in contrast to our proposal, these methods fail to scale-up for longer task sequences.

In the second category, we have replay/rehearsal based methods that attempt to simultaneously retrain the continual learning model over the previous tasks data and the current task data to avoid catastrophic forgetting \citep{lopez2017gradient, riemer2018learning, rios2018closed, shin2017continual, zhang2019prototype, Rebuffi2016iCaRLIC, Chaudhry2019OnTE, AGEM}. Obtaining samples or knowledge from previous tasks is usually performed in two different ways: (i) adding a generative model to the continual learning model and producing samples of earlier tasks \cite{shin2017continual}, (ii) augmenting a memory unit to the continual learning model and accumulating a small subset of raw input data \cite{Chaudhry2019OnTE} or gradient parameters  \cite{AGEM}. Recently, \cite{saha2021gradient} propose a memory-based method where a new task is learned by taking gradient steps in the orthogonal direction to the gradient subspaces marked as crucial for previous past tasks. The method employs SVD after learning each task to find the crucial subspaces and stores them in a memory. 
In our proposed method, rather than using a memory unit, we augment the kernel continual learning model by \cite{derakhshani2021kernel} with a conditional generative model to enable generation of samples for each task.

The third category covers architecture-based methods \citep{rusu2016progressive, yoon2018lifelong, Jerfel2018ReconcilingMA, li2019learn, wortsman2020supermasks, Packnet}. 
These methods aim to directly minimize the task interference problem by either pruning and model expansion \cite{Packnet}, or allocating a set of new parameters when observing a new task \cite{rusu2016progressive}, or by partioning a neural network into several sub-networks using gating mechanism \cite{wortsman2020supermasks}. 
Recently, \cite{pmlr-v139-kumar21a} presents a Bayesian framework to learn the structure of deep neural networks by unifying the variational Bayes based regularization and architecture based methods. This method supports knowledge transfer among tasks by overlapping sparse subsets of weights learned by different tasks. 
Using an expectation maximization method, \cite{pmlr-v139-lee21a} introduces a transfer mechanism that selectively chooses the transfer architecture configuration for each task. For each task, this would allow the method to dynamically select which layers to transfer and which to keep as task-specific.
\cite{veniat2021efficient} introduce a compositional neural architecture for continual learning, where each module in the network represents an atomic skill and can be combined to solve a certain task. Similarly, our generative kernel continual learning adopts a gating mechanism in the decoder of the conditional variational auto-encoder to minimize task interference.

\section{Conclusion}
\label{lbl:conclusion}
In this paper, we introduce generative kernel continual learning, a memory-less variant of kernel continual learning that replaces the episodic memory with a generative model based on variational auto-encoders. We further introduce supervised contrastive regularization, which enables our model to generate even more discriminative samples for better classification performance. We conduct extensive experiments on three benchmarks for continual learning. Our experiments highlight the effectiveness of generative kernel continual learning. First, it is shown that synergizing the strengths of generative models and kernels leads to remove the dependence on an explicit memory while being able to tackle catastrophic forgetting and task interference. Second, it is  demonstrated that adding a supervised contrastive loss into the generative modelling increases the discriminability of the generated latent representations, improving the model’s performance in terms of accuracy and forgetting. Moreover, we show that our generative kernel continual learning already achieves state-of-the-art performance on all benchmarks with a simple linear kernel.

\section*{Acknowledgements}
This work is financially supported by the Inception Institute of Artificial Intelligence, the University of Amsterdam and the allowance Top consortia for Knowledge and Innovation (TKIs) from the Netherlands Ministry of Economic Affairs and Climate Policy.
\newpage
\section*{Ethics Statement}
Being able to adapt to non-stationary data distributions and continuously changing environments, our method has potential inherent impact in the applications that often encounter dynamic environments in practice, e.g., medical imaging, astronomical imaging, and autonomous driving.
Accordingly, our method would also potentially face some negative social impacts accompanying with applications, e.g., lack of fairness with the model trained by incomplete data, legal compliance, and the privacy of patients in medical imaging.

\section*{Reproducibility Statement}
We refer to Section \ref{lbl:experiments_benchmarks} for detailed information on benchmarks, metrics, and the implementation of generative kernel continual learning in terms of architecture and training. In addition, we refer to Appendix \ref{lbl:hyper} for the list of all hyperparameters used to train generative kernel continual learning. We will further open-source all code, scripts to reproduce the experiments with the exact hyperparameters, and scripts to calculate the evaluation results at: \url{https://github.com/<redacted>/<redacted>}.

\bibliography{main}
\bibliographystyle{iclr2022_conference}

\newpage
\appendix
\section{Appendix}
In section \ref{lbl:hyper}, we report all hyperparameters used to reproduce the results in Table \ref{tab:compare-20tasks} and Figure \ref{fig:compare_all}. In section \ref{lbl:kernel_network}, we study the performance of generative kernel continual learning with and without kernel network $f_\gamma$. Finally, In section \ref{lbl:feat_vis}, we visualize the internal representation of conditional variational auto-encoder, which is used as the input for the kernel network $f_\gamma$, with and without contrastive regularization term. 

\subsection{Hyperparamters}
\label{lbl:hyper}
In Table \ref{tab:f_gamma_hyper} and \ref{tab:gen_hyper}, we report all hyperparameters used to generate results in Table \ref{tab:compare-20tasks} and Figure \ref{fig:compare_all} in the main paper.

\begin{table}[!htbp]
\centering
\caption{Hyperparameters used to train the kernel network $f_\gamma$.}
\vspace{2mm}
\label{tab:f_gamma_hyper}
\resizebox{.7\columnwidth}{!}{%
\begin{tabular}{@{}lccc@{}}
\toprule
& \textbf{RotatedMNIST} & \textbf{PermutedMNIST} & \textbf{SplitCIFAR100} \\ 
\midrule
Batch size         & 10  & 10  & 10     \\
Learning rate (LR) & 0.1 & 0.1 & 0.3    \\
LR decay factor    & 0.8 & 0.8 & 0.95   \\
Momentum           & 0.8 & 0.8 & 0.4    \\
Dropout            & 0.1 & 0.1 & 0.0    \\
Coreset size       & 20  & 20  & 20     \\
Kerenl type        & Linear & Linear & Linear \\
Optimizer          & SGD & SGD & SGD \\

\bottomrule
\end{tabular}%
}
\end{table}

\begin{table}[h]
\centering
\caption{Hyperparameters used to train the conditioanl variational auto-encoder.}
 \vspace{2mm}
\label{tab:gen_hyper}
\resizebox{.7\columnwidth}{!}{%
\begin{tabular}{@{}lccc@{}}
\toprule
& \textbf{RotatedMNIST} & \textbf{PermutedMNIST} & \textbf{SplitCIFAR100} \\ 
\midrule
Learning Rate       & 0.001 & 0.001 & 0.001 \\
Batch   size          & 512 & 512 & 512     \\
Replay size          & 512 & 512 & 512      \\
Number of iteration & 2000 & 2000 & 300     \\
Optimizer           & Adam & Adam & Adam    \\
Temperature ($\tau$)              & 0.08 & 0.08 & 0.04 \\
\bottomrule
\end{tabular}%
}
\end{table}

\subsection{Influence of Kernel network}
\label{lbl:kernel_network}
In this section, we study the performance of generative kernel continual learning with and without kernel network $f_\gamma$. This network maps the internal representation of conditional
\begin{table}
\caption{\textbf{Influence of kernel network $f_\gamma$}. Average accuracy of our proposed method with and without kernel network over 20 sequential  tasks for five different random seeds on SplitCIFAR100. Incorporating kernel network into the conditional variational auto-encoder increases average accuracy considerably.}
\label{tab:ablation_kernel_network}
\begin{center}
\begin{tabular}{@{}lc@{}}
\toprule
& \textbf{SplitCIFAR100} \\ 
\midrule
with kernel network          & 72.79  \\
without kernel network         & 59.68  \\
\bottomrule
\end{tabular}
\end{center}
\end{table}
variational auto-encoder to another new space, and upon this new space, we construct each task's non-parametric classifier. For fair comparison, in both experiments, we exploit same hyperparameters and architecture as provided in Tables \ref{tab:f_gamma_hyper} and \ref{tab:gen_hyper}. Moreover, we train both models for $20$ sequential tasks on SplitCIFAR100 benchmark, and report the average accuracy over $5$ different random seeds. Results are presented in Table \ref{tab:ablation_kernel_network}. As it is shown, augmenting generative kernel continual learning with kernel network $f_\gamma$ enhances the model performance by $12\%$.

\subsection{Feature Space Visualization}
\label{lbl:feat_vis}
\begin{figure*}[h]
\centering
\includegraphics[width=1\textwidth]{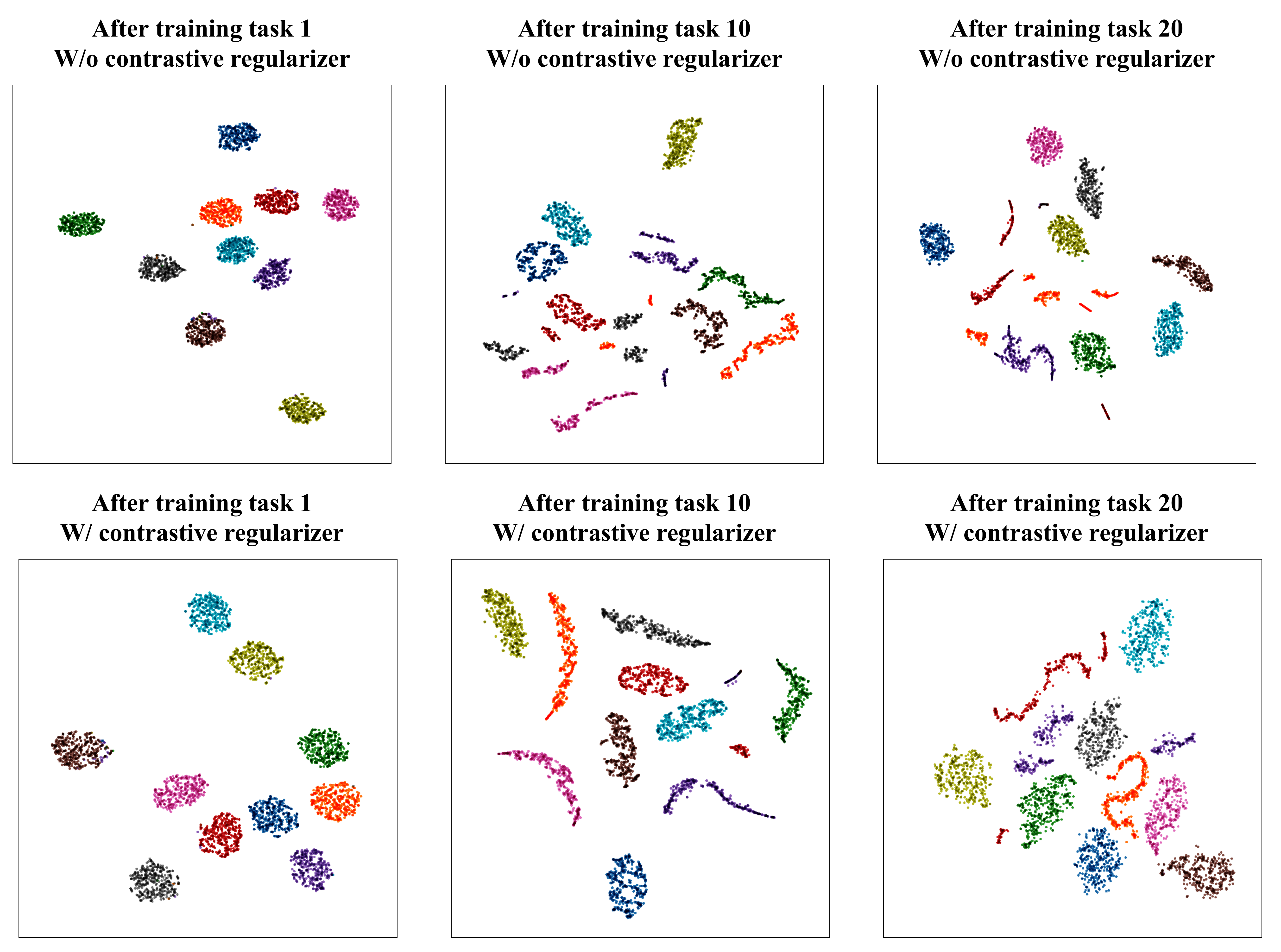}
\caption{\textbf{Coreset internal representation}. In this figure, we visualize the internal representation of the encoder network of conditional variational auto-encoder of the first task when we observe 20 tasks with (w/) and without (w/o) the contrastive regularization term. As shown, the regularization term allows our proposed method to obtain better concentrated features.}
\label{fig:tsne_coreset}
\end{figure*}

\begin{figure*}[h]
\centering
\includegraphics[width=1\textwidth]{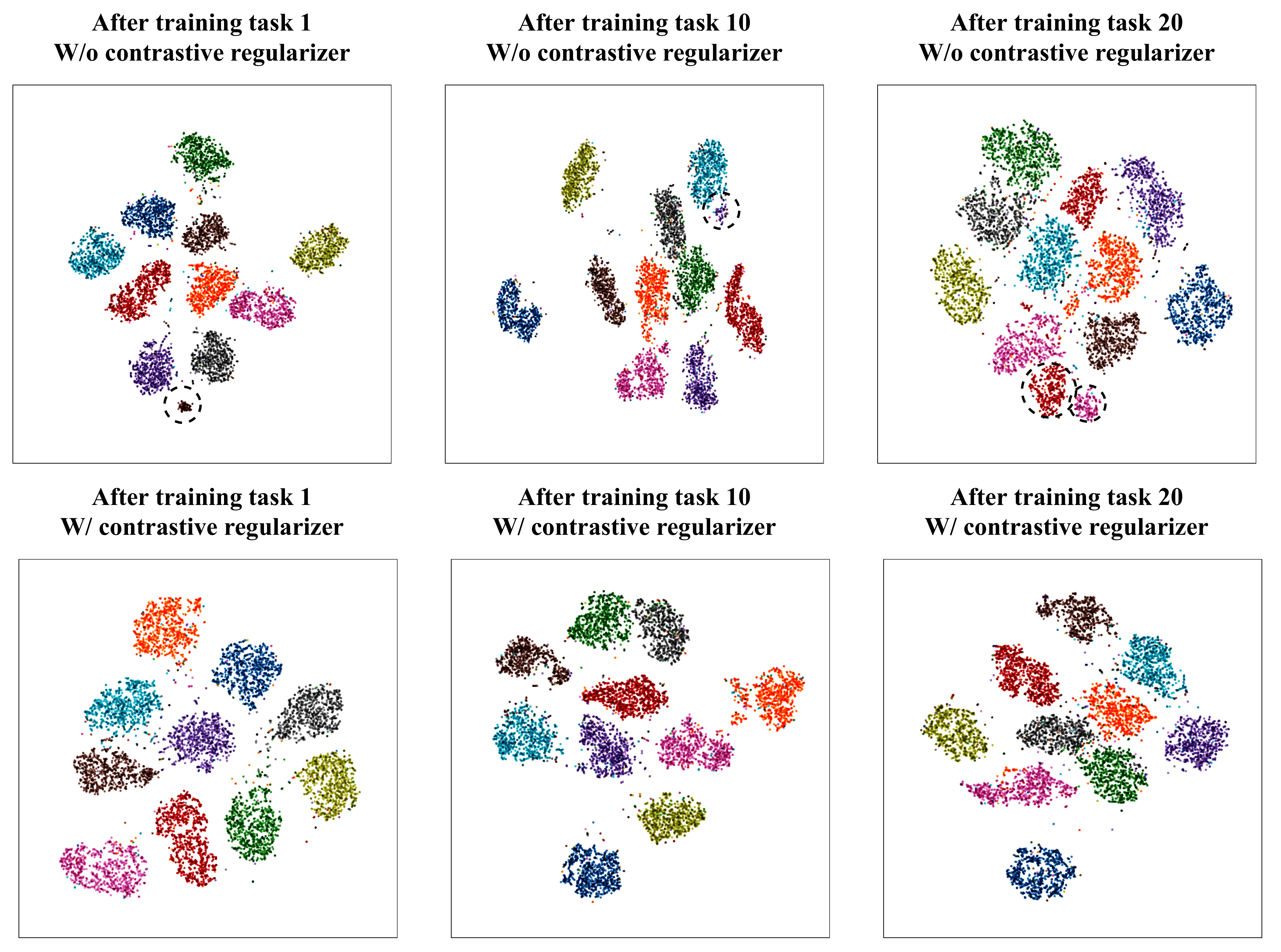}
\caption{\textbf{Test dataset internal representation}. In this figure, we visualize the internal representation of the encoder network of conditional variational auto-encoder of the first task when we observe 20 tasks with (w/) and without (w/o) the contrastive regularization term. As shown, the regularization term allows our proposed method to obtain better concentrated features.}
\label{fig:tsne_test}
\end{figure*}

To highlight the benefit of the supervised contrastive regularization term in enhancing the discriminability of the generated internal representations, we further visualize the internal representation of the encoder network $q_\phi$ on the generated coreset of task $1$ in Figure \ref{fig:tsne_coreset} where we train the generative kernel continual learning model over 20 sequential tasks. In this figure, the first row shows the scenario where we train generative kernel continual learning without supervised contrastive regularization term while the second row is the case where we exploit supervised contrastive regularization. Comparing these two scenarios shows that the regularization term leads to improve the discriminability of internal representation of the conditional variational auto-encoder. To explore more, we perform similar experiments over the test dataset in Figure \ref{fig:tsne_test}. Same conclusion as coreset is inferred.  

\end{document}